# EfficientNet-Based Multi-Class Detection of Real, Deepfake, and Plastic Surgery Faces

Li Kun, Milena Radenkovic
{alykl17, milena.radenkovic} @ nottingham.ac.uk
School of Computer Science, The University of Nottingham, Nottingham NG8 1BB, UK

**Abstract**

Currently, deep learning has been utilised to tackle several difficulties in our everyday lives. It not only exhibits progress in computer vision but also constitutes the foundation for several revolutionary technologies. Nonetheless, similar to all phenomena, the use of deep learning in diverse domains has produced a multifaceted interaction of advantages and disadvantages for human society. Deepfake technology has advanced, significantly impacting social life. However, developments in this technology can affect privacy, the reputations of prominent personalities, and national security via software development. It can produce indistinguishable counterfeit photographs and films, potentially impairing the functionality of facial recognition systems, so presenting a significant risk.

The improper application of deepfake technology produces several detrimental effects on society. Face-swapping programs mislead users by altering persons' appearances or expressions to fulfil particular aims or to appropriate personal information. Deepfake technology permeates daily life through such techniques. Certain individuals endeavour to sabotage election campaigns or subvert prominent political figures by creating deceptive pictures to influence public perception, causing significant harm to a nation's political and economic structure.

## 1.Introduction

Deepfake image and video technology has gained a lot of attention due to the quick advancement of artificial intelligence technology. The technology primarily forges human faces or videos using deep learning models like Generative Adversarial Networks (GAN). The resulting images and videos are incredibly lifelike and hard to tell apart from normal human observation (Patel et al., 2023).

Deepfake has been extensively utilised in negative contexts in recent years, including identity theft, online fraud, revenge pornography, fake news, and political manipulation (Ahmed et al., 2022). The authenticity of online information is seriously threatened by the quick spread of Deepfake technology, which may result in major social, political, and even financial issues (Raza et al., 2022). Softwares related to deepfake such as: face app and fake app are easy to use and low cost, any ordinary person can get their desired image or video processed by Deepfake through simple operation (Guera & Delp, 2018). According to Sensity, more than 96 per cent of Deepfake content is obscene. in 2019, cybercriminals tricked their CEOs into completing their crimes by transferring $243,000 to their bank accounts through fake, fake audio content.

Therefore, it is imperative that we use advanced forgery detection tools to control and detect the rise in crime rates associated with Deepfake (Raza et al., 2022).

## 2. MTCNN Feature Extraction

MTCNN is a prevalent feature extraction technique, recognised as a multi-task convolutional neural network(Zhang et al., 2020).

An image pyramid is initially created by proportionally scaling the image with a scaling factor, usually set at 0.709. The image undergoes several scaling operations, with each scaled version constituting a tier of the pyramid structure. Scaling ceases when each dimension of the image is less than 12 pixels.

This stair-step scaling pattern delineates the image pyramid, intended to identify facial characteristics of diverse dimensions.

P-Net predominantly offers proposed bounding box positions. Images processed in the preceding step are subjected to the FCN fully connected layer for preliminary feature extraction and bounding box annotation. Following the box selection process, two parameters are produced: classifier and bbox_regress. These denote the confidence level of the box at this grid point and the positioning information of the box within the grid point. Furthermore, we eliminate superfluous boxes by non-maximum suppression. For each region, we choose the box with the best score and eliminate others that significantly overlap. By removing superfluous boxes and preserving more precise ones, we enhance the prediction windows for more feature extraction.

R-Net is a residual neural network architecture. In contrast to P-Net, R-Net has an extra fully linked layer, enforcing more stringent limitations on input data. Prediction windows analysed by P-Net undergo scaling adjustments and score calculations. The modified boxes more accurately correspond to the dimensions of the face in the image. The output of R-Net, similar to that of P-Net, functions to reduce the number of non-face bounding boxes.

The workflow of O-Net resembles that of R-Net. It is a more intricate convolutional neural network that incorporates an additional neural layer in comparison to R-Net. This layer conducts additional edge detection and verifies the existence of a face. Enhance by using a neural layer for improved boundary selection and face presence identification. O-Net processes images to accurately identify five key feature points: left eye, right eye, nose, left corner of the mouth, and right corner of the mouth. O-Net generates three outputs: classifier, bbox_regress, and landmark_regress (the five face landmarks).

The MTCNN feature extraction technique consistently enhances the recognition of face regions. It refines initial crude bounding boxes into more precise ones through many rounds of filtering and detection. It additionally modifies the selection ratio and confirms whether the chosen regions encompass faces. This repeated detection and refining procedure extracts five face landmarks, so completing the facial feature extraction.

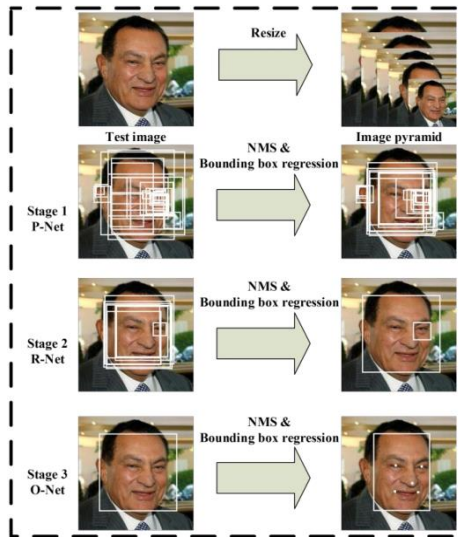
Fig. 1. Pipeline of MTCNN(Zhang et al., 2016)

### 3. The Azure Face API executes facial recognition

The Azure Face API is a complimentary API. The computer vision services on Azure are both broadly applicable and highly precise, rendering them a suitable choice for facial recognition. Computer vision is a component of Microsoft's cognitive services, consisting of sophisticated algorithms that developers can utilise to obtain processed information. Images can be uploaded directly or studied using designated image URLs in the context of image processing. The image processing algorithms include several functions, such as identifying improper content, classifying photos, detecting faces, characterising image details, recognising objects, and calculating the age of humans in photographs. It has a broad spectrum of truly dependable functionalities. This article employs the face detection capability of these image processing methods.

Certain limitations are inherently imposed when utilising this API for photos. The image format must be either JPEG, PNG, GIF, or BMP. Therefore, any uploaded video needs undergo processing by removing frames and converting them into picture format prior to facial feature extraction. Additionally, the image file size must not surpass 4MB, and there are limitations on the image dimensions. The image resolution must exceed 50×50 pixels. These represent the prerequisites for image uploads.

### 4. EfficientNet

In computer vision, as well as in machine learning and deep learning, three prevalent strategies are utilised to augment project accuracy: developing a completely new algorithm; formulating a groundbreaking algorithm that transforms efficiency and precision; or enhancing the performance of existing network models. EfficientNet improves total network performance in three aspects: depth, width, and input image resolution (Pipit Utami et al., 2022).

Augmenting the depth of convolutional networks facilitates the acquisition of more intricate and nuanced characteristics while providing substantial scalability. The ResNet model introduced by He Kai's team in 2015 rendered the training of networks with hundreds or even thousands of layers practicable. This advancement created new opportunities for deep learning. The universal approximation theorem posits that a single-layer feedforward network may represent any function, provided there is adequate capacity. Nevertheless, such a layer may become overly large, perhaps resulting in overfitting. This requires more complex network architectures. Furthermore, several networks persist in enhancing convolutional topologies. The VGG network improves performance by incrementally adding layers to enhance depth. However, above a specific threshold, additional increases in layer count achieve saturation, rendering substantial performance improvements challenging. At this juncture, alterations to other dimensions become imperative .

Augmenting the width of the neural network entails increasing the quantity of convolutional kernels, particularly the number of output channels. This method significantly improves the efficacy of smaller networks by facilitating the extraction of more intricate visual information. Nevertheless, networks that are overly broad but shallow encounter difficulties in capturing high-level information. Analogous to augmenting depth, expanding the network beyond a specific threshold results in saturation, wherein additional enhancements in accuracy yield progressively diminishing returns.

Augmenting the resolution of a neural network can also improve its efficacy. To enhance image clarity, the network may extract more intricate elements, resulting in superior performance.

Optimising the depth, breadth, and resolution of a convolutional neural network is essential for enhancing its performance. Modifying the depth, width, or resolution of a neural network can enhance accuracy; but, for bigger models, the improvement in accuracy becomes negligible.

$$Y_i = F_i(x_i)\langle H_i, W_i, C_i \rangle \tag{4,1}$$

$$N = F_k \odot \ldots \odot F_2 \odot F_1(x1) = \underset{j=1}{\overset{k}{\odot}} F_j(x_1) \tag{4,2}$$

$$N = \underset{i=1\ldots s}{\odot} F_i^{L_i}(X_{\langle H_i, W_i, C_i \rangle}) \tag{4,3}$$

$$\underset{d,w,r}{\max} Accuracy(N(d,w,r)) \tag{4,4}$$

$$s.t. N(d,w,r) = \underset{i=1\ldots s}{\odot} \hat{F}_i^{d \cdot \hat{L}_i}(X_{\langle r \cdot \hat{H}_i, r \cdot \hat{W}_i, w \cdot \hat{C}_i \rangle}) \tag{4,5}$$

$$Memory(N) \leq target\_memory \tag{4,6}$$

$$FLOPS(N) \leq target\_flops \tag{4,7}$$

Thus, achieving equilibrium across each dimension and scaling proportionately represents the function executed by the EfficientNet network. expanding network depth alone leads to rapid performance saturation; expanding network width alone also results in swift performance saturation; similarly, resolution will eventually attain saturation. Thus, concurrently altering all three dimensions produces enhanced outcomes.

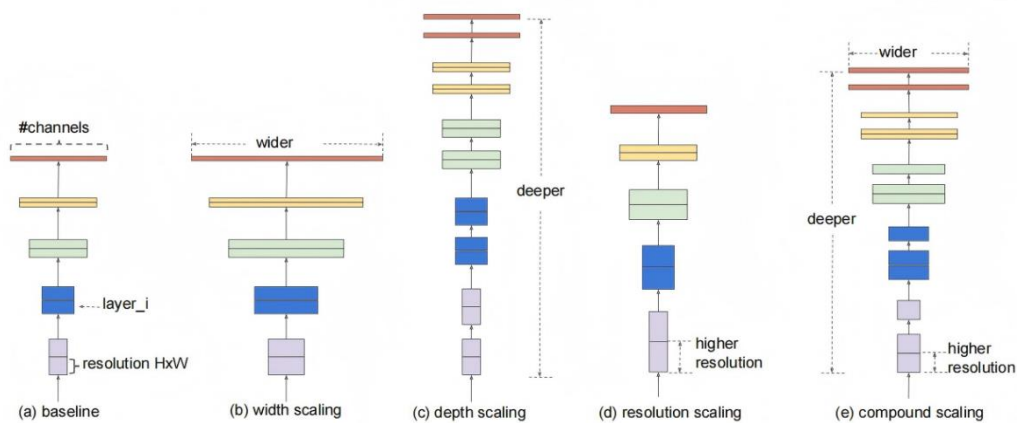

Fig.2 Characteristics of the EfficientNet Architecture(Tan & Le, 2019)

## 5. ResNet-50

ResNet is a neural network based on convolutional neural networks (CNNs). ResNet is a straightforward and effective network architecture extensively utilised in detection, segmentation, recognition, and various other domains(He et al., n.d.).

ResNet consists of five steps. Stage 0 presents a simple architecture functioning as an input preprocessing unit, but the following four stages consist of Bottleneck layers, demonstrating significant structural resemblance. The phases comprise 3, 4, 6, and 3 Bottleneck layers, respectively.

In Stage 0, the input image undergoes sequential processing through a convolutional layer, a Batch Normalisation (BN) layer, and a ReLU activation function. This entails a single convolutional operation succeeded by a max pooling operation. The BN layer principle is to normalise the input distribution from each hidden layer neurone, which progressively aligns with non-linear functions and nears the saturation area of the output range, to a mean of 0 and a variance of 1, thereby approximating a standard normal distribution. This method expedites the convergence rate of the network, alleviates the vanishing gradient issue, and improves the network's generalisation ability to a typical normal distribution with a mean of 0 and a variance of 1. This expedites network convergence, alleviates gradient vanishing, and improves the network's generalisation capacity.

Residual blocks are integrated into Stages 1, 2, 3, and 4. Each residual block consists of a convolutional layer, a normalisation layer, an activation function (ReLU), and a residual feedback mechanism, with a variable number of convolutional kernels. The number of residual blocks varies over phases, as does the quantity of convolutional kernels within them.

The concluding phase implements average pooling, succeeded by a rectification layer, and culminates in a fully linked layer that produces probabilities.

The residual network improves the efficacy of each layer without augmenting the overall network complexity, mitigating gradient explosion and facilitating end-to-end backpropagation training.

## 6. VGG

VGG was introduced by the Visual Geometry Group at Oxford University, as shown by the network's nomenclature derived from the initials of its founding group. This design was unveiled at the 2014 ImageNet Large Scale Visual Recognition Challenge. VGG utilises a more profound network architecture, illustrating that augmenting network depth can markedly affect its overall performance. VGG comprises several topologies, notably VGG16 and VGG19, which vary solely in network depth without significant structural differences. This study utilises the VGG16 model as the benchmark for comparison analyses(Simonyan & Zisserman, 2015).

The primary benefit of the VGG architecture is its sole reliance on 3×3 convolutional kernels. The advantages of utilising small convolutional kernels include the substitution of larger kernels, which decreases the total number of network parameters. Furthermore, smaller kernels more effectively maintain picture characteristics, thereby improving overall network accuracy. The significant decrease in parameters enables numerous convolutional layers with limited receptive fields to substitute for one layer with an extensive receptive field, thus enhancing the network's nonlinear expressive capacity. The model utilises 2×2 pooling kernels. Convolutional kernels enhance channel counts, however pooling diminishes breadth and height, allowing for the development of deeper and wider models while controlling computational expansion. Moreover, by replacing fully connected layers with convolutional kernels, VGG permits inputs of variable dimensions.

## 7. Dataset

In this paper, two publicly available datasets are used to train, test and validate the performance of the Deepfake detection model. These datasets are sourced from credible research institutions or data platforms and contain real and fake face images as well as before and after cosmetic surgery photos, which help to build a detection mechanism against deepfake techniques.

1. HDA Plastic Surgery Face Database

This dataset is provided by the Hochschule Darmstadt (University of Applied Sciences, Darmstadt, Germany), and the original URL is: https://dasec.h-da.de/research/biometrics/hda-plastic-surgery-face-database/. The dataset contains images of faces before and after undergoing plastic surgery, and is used to study the impact of facial changes on biometric recognition algorithms. Although this dataset is not specifically designed for Deepfake images, it contains images of face changes that can be used to test whether a deepfake detection system can recognize physically surgically altered faces, and is therefore used as an extended dataset in this study to improve the robustness of the model to face deformations.

2. DeeperForensics-1.0

This research utilised the DeeperForensics-1.0 dataset (Jiang et al., 2020), a comprehensive and high-caliber benchmark specifically developed for the identification of authentic deepfakes. The collection contains more than 60,000 movies (about 17.6 million frames) showcasing 100 performers from various demographic backgrounds, recorded under controlled variations in posture, facial expressions, and lighting conditions. To preprocess the data, extract frames from 300 videos and store them as pictures. To fulfil later processing needs, crop the photos. Videos with a width under 300 pixels will be resized to double their original dimensions; videos with widths ranging from 300 to 1000 pixels will retain their original dimensions; videos with widths between 1000 and 1900 pixels will be resized to half their original dimensions; images exceeding 1900 pixels in width will be resized to one-third of their original dimensions.

Subsequently, for face feature recognition, we utilised both the MTCNN feature extraction technique and Azure's computer vision facial recognition capabilities to extract features. The outcomes of these feature extractions were preserved individually as independent images. When many items are present in a single video, we catalogue them individually, thereby enhancing our dataset. We subsequently applied a 30% margin to either side of the identified bounding boxes and gathered the facial photographs with a 95% confidence threshold.

## 8. Deepfake Detection Based on the EfficietNet-B4 Model

Activation function: The ReLU function is a piecewise linear function. If the input data is positive, it outputs the data directly as the result. If the input is zero or negative, it outputs zero. The formula is as follows:

$$\mathrm{Re}\,LU(x) = \max(0, x) \qquad (8,1)$$

One popular activation function in neural networks is the sigmoid function. It generates the likelihood of forgery detection in this project, acting as the output layer. The following is the formula:

$$Sigmoid(x) = \frac{1}{1+e^{-x}} \qquad (8,2)$$

This function's benefit is that it is easy to implement, produces results in the [0,1] range that may be displayed as probability outputs, and is not easily affected by noisy data. The incidence of gradient saturation is a disadvantage.

Utilise the Adam optimiser. Adam, formally referred to as adaptive moment estimation, is commonly known as adaptive moment estimation. The implementation is simple and highly efficient, typically demonstrating itself as a good option.

The model structure is as follows:

| Layer(type) | Output shape | Param |
|---|---|---|
| Input (InputLayer) | [None, 380, 380, 3] | 0 |
| EfficientNet-B4 (Functional) | (None, 1792) | 17673820 |
| Dropout (Dropout) | (None, 1792) | 0 |
| Dense (Dense) | (None, 3) | 5379 |
| Total params: 17679199 Trainable params: 5379 Non-trainable params: 17673820 | | |

The dataset has been segmented into several files, categorising all photos into training, validation, and test sets. The segmented findings are presented in the table below. The training, validation, and test datasets are distributed in an 7:2:1 proportion.

|  | count |
|---|---|
| train | 144866 |
| validation | 39428 |
| test | 11446 |

The pertinent data for the model's initial nine iterations is displayed in the table below, A total of fifteen iterations were traversed, with only nine displayed.:

| Epoch | Train Loss | Train Acc | Val Loss | Val Acc |
|---|---|---|---|---|
| 1 | 0.1486 | 0.9397 | 0.1252 | 0.9579 |
| 2 | 0.0916 | 0.9622 | 0.0740 | 0.9726 |
| 3 | 0.0810 | 0.9664 | 0.0553 | 0.9802 |
| 4 | 0.0761 | 0.9686 | 0.0852 | 0.9741 |
| 5 | 0.0711 | 0.9702 | 0.0670 | 0.9772 |
| 6 | 0.0693 | 0.9715 | 0.0459 | 0.9852 |
| 7 | 0.0658 | 0.9726 | 0.0962 | 0.9711 |
| 8 | 0.0651 | 0.9724 | 0.0431 | 0.9856 |
| 9 | 0.0621 | 0.9737 | 0.0508 | 0.9834 |

The loss and accuracy for the test set and validation set are functions of epochs, as depicted in the figure below:

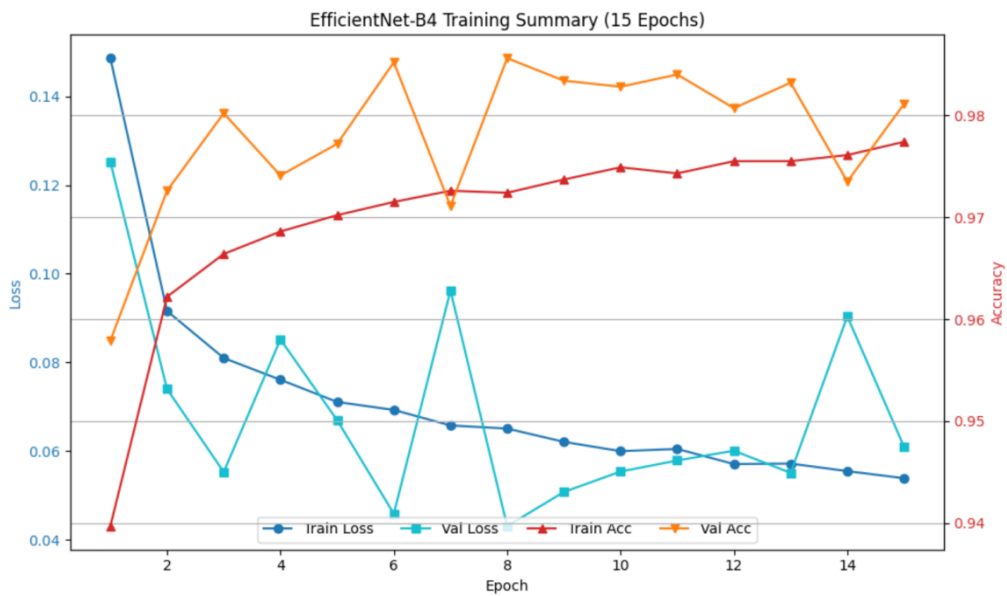

Fig.3 EfficientNet-B4 Model Loss Rate and Accuracy Variation Chart

## 9. Deepfake Detection Based on the ResNet-50 Model

| Layer(type） | Output shape | Param |
|---|---|---|
| Input_12(InputLayer) | [(None,128,128,3)] | 0 |
| Sequential_2(Sequential) | (None,128,128,3) | 0 |
| Tf._operaytors_.getitem_3(SlicingOpLambda) | (None,128,128,3) | 0 |
| Tf.nn.bias_add_3(TFOplambda) | (None,128,128,3) | 0 |
| Resnet50(Functional) | (None,4,4,2048) | 23587712 |
| Global_average_polling2d_3(GlobalAveragePooling2D) | (None,2048) | 0 |
| Dense_3(Dense) | (None,2) | 4098 |
| Total params:23591810 Trainable params:4098 Non-trainable params:23587712 | | |

The network configuration employs a callbacks mechanism with an epoch set to 100. Due to the model fitting progressing too slowly, the callbacks mechanism was not triggered. Relevant data per 10 iterations is presented in the table below:

| Epoch | Train Loss | Train Acc | Val Loss | Val Acc |
| --- | --- | --- | --- | --- |
| 1 | 0.6778 | 0.572 | 0.6616 | 0.594 |
| 10 | 0.5039 | 0.749 | 0.5008 | 0.779 |
| 20 | 0.4722 | 0.777 | 0.4528 | 0.783 |
| 30 | 0.4822 | 0.774 | 0.4391 | 0.812 |
| 40 | 0.4466 | 0.788 | 0.4334 | 0.785 |
| 50 | 0.4313 | 0.794 | 0.4413 | 0.793 |
| 60 | 0.2199 | 0.906 | 0.1533 | 0.926 |
| 70 | 0.1931 | 0.918 | 0.4525 | 0.811 |
| 80 | 0.1923 | 0.917 | 0.1314 | 0.949 |
| 90 | 0.1776 | 0.928 | 0.0909 | 0.967 |
| 100 | 0.1519 | 0.934 | 0.1273 | 0.938 |

The changes in training loss, training accuracy, validation loss, and validation accuracy with respect to the number of epochs are illustrated in the figure below:

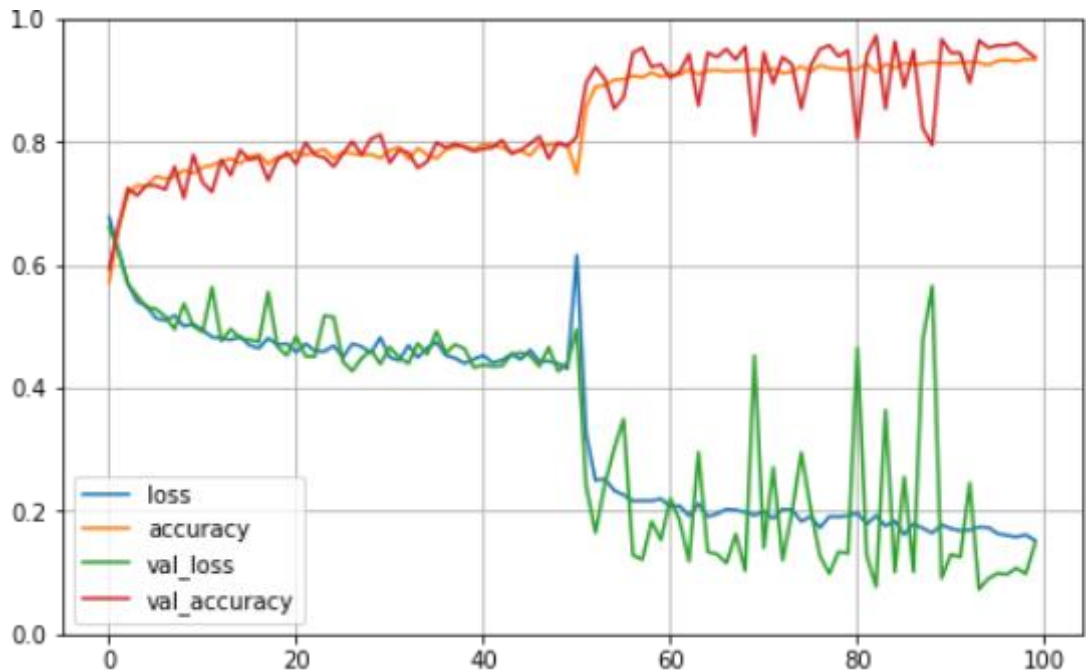

Fig.4 ResNet-50 Model Loss Rate and Accuracy Variation Chart

## 10. Deepfake Detection Based on the VGG-16 Model

| Layer(type） | Output shape | Param |
|---|---|---|
| Input_8(InputLayer) | [(None,128,128,3)] | 0 |
| Tf._operators_.getitem_2(SlicingOpLambda) | (None,128,128,3) | 0 |
| Tf.nn.bias_add_2(TFOpLambda) | (None,128,128,3) | 0 |
| Vgg16(Functional) | (None,4,4,512) | 14714688 |
| Global_average_pooling2d_2(GlobalAveragePooling2D) | (None,512) | 0 |
| Dense_2(Dense) | (None.2) | 1026 |
| Total params:14715714 | | |
| Trainable params:1026 | | |
| Non-trainable params:14714688 | | |

The network configuration employs a callbacks mechanism with an epoch set to 100. Due to the model fitting progressing too slowly, the callbacks mechanism was not triggered. Relevant data per 10 iterations is presented in the table below:

| Epoch | Train Loss | Train Acc | Val Loss | Val Acc |
|---|---|---|---|---|
| 1 | 0.678 | 0.596 | 0.658 | 0.584 |
| 10 | 0.589 | 0.684 | 0.604 | 0.713 |
| 20 | 0.564 | 0.714 | 0.571 | 0.713 |
| 30 | 0.562 | 0.712 | 0.567 | 0.713 |
| 40 | 0.541 | 0.721 | 0.554 | 0.746 |
| 50 | 0.517 | 0.741 | 0.562 | 0.715 |
| 60 | 0.248 | 0.896 | 0.163 | 0.945 |
| 70 | 0.166 | 0.931 | 0.138 | 0.961 |
| 80 | 0.109 | 0.951 | 0.045 | 0.982 |
| 90 | 0.097 | 0.957 | 0.046 | 0.969 |
| 100 | 0.088 | 0.960 | 0.052 | 0.973 |

The changes in training loss, training accuracy, validation loss, and validation accuracy with respect to the number of epochs are illustrated in the figure below:

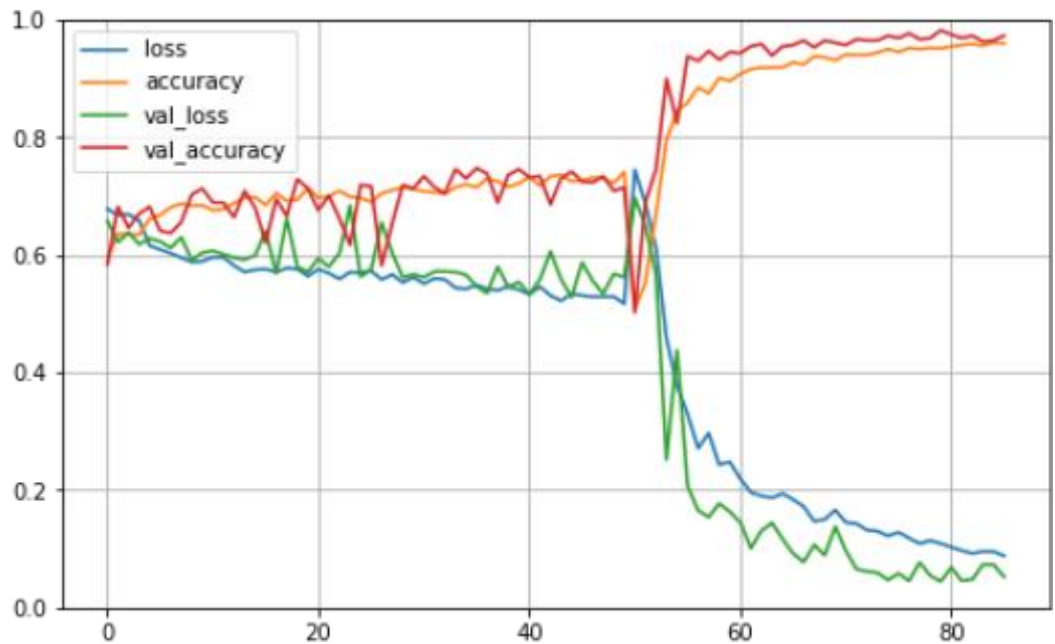

Fig.5 VGG-16 Model Loss Rate and Accuracy Variation Chart

## 11.Result

The EfficietNet-B4 model achieves 97% accuracy after just 5 epochs; the ResNet-50 model, after 100 training rounds, achieves less than 94% accuracy; the VGG-16 model, after 100 training rounds, achieves 97% accuracy. Concurrently, we conducted validation testing on the model using 11,446 images. The confusion matrix for the EfficietNet-B4 model's test results on the 11,446-image test dataset is shown below:

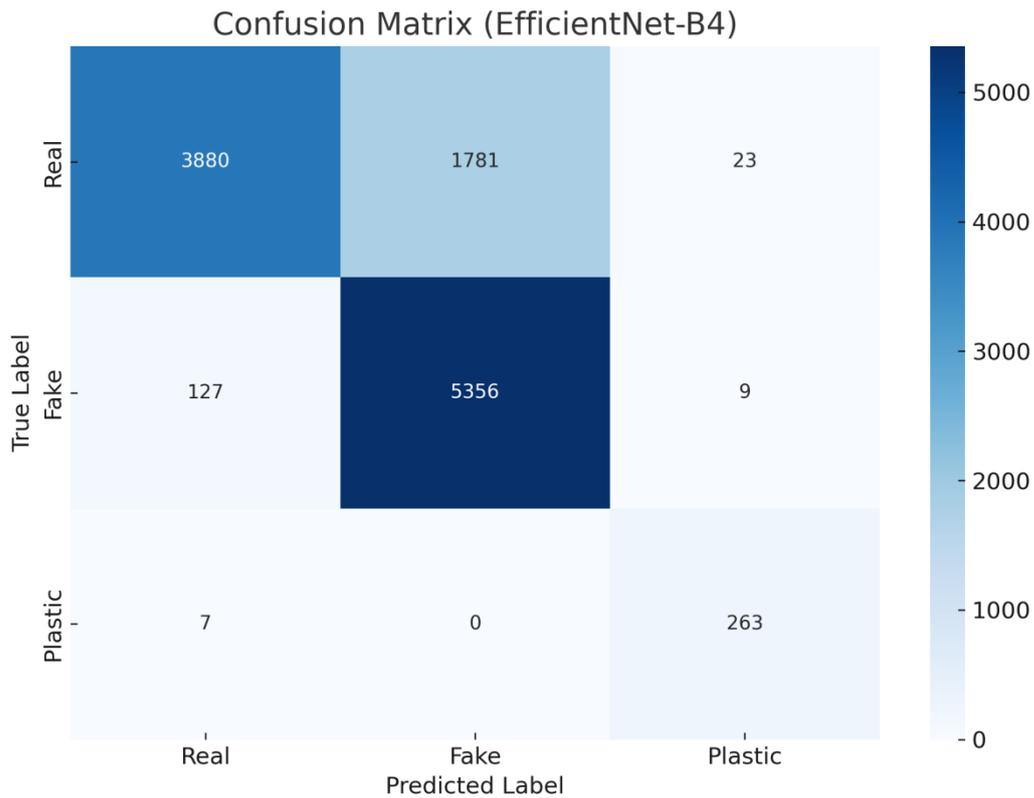

Fig.6 Confusion Matrix for the EfficietNet-B4 Model on the 11,446 Test Dataset

The confusion matrix reveals the model's generalisation capability. As shown in the table below:

|  | Precison | Recall | fi-score | Support |
|---|---|---|---|---|
| Real | 0.9666 | 0.6826 | 0.8002 | 5684 |
| Fake | 0.7505 | 0.9752 | 0.8482 | 5492 |
| Plastic | 0.8915 | 0.9741 | 0.9310 | 270 |
| Accuracy |  |  | 0.8299 | 11446 |
| Macro avg | 0.8695 | 0.8773 | 0.8598 | 11446 |
| Weighted avg | 0.8611 | 0.8299 | 0.8263 | 11446 |

## 12.Front-end System Design

PyQt5 was utilised as the basis for developing the system's frontend. This framework offers several advantages: a diverse selection of controls for comprehensive GUI application development; exceptional cross-platform compatibility across multiple operating systems; support for Qt's visual designer for graphical interface design; and an extensive library containing 620 classes and over 6,000 functions and methods, adequate for implementing all necessary functionalities. Three modules were employed: QtCore, QtGui, and QtWidgets. Their corresponding functions are detailed in the table below:

| PyQt5 | Function |
|---|---|
| QtCore | QtCore encompasses processing time, data types, object types, files, and directories required during compilation. It forms the foundation for all other Qt modules and constitutes the core non-GUI functionality within the package. |
| QtGui | QtGui encompasses functionality pertaining to window system classes, fundamental graphics, and font types. |
| QtWidgets | QtWidgets provides a set of UI element classes for creating classic desktop-style user interfaces. |

The front-end facial recognition feature is executed through the `face_recognition()` method. Before starting with verification, it is imperative to detect faces in the provided photographs, as they may not contain any. `face_recognition()` is a resilient, user-friendly, and readily accessible library. Its superior interoperability renders it particularly advantageous for front-end facial recognition applications. Upon executing the face_recognition() function to analyse an image, the system displays the message: 'No face found in the uploaded image.' Kindly re-upload if facial features are not detected. If a face is detected, the system will crop the image and present it within a sample frame. The data image is subsequently input into the trained model for authenticity checking. The resultant detection conclusion is subsequently presented on the front-end display interface.

The interface includes image submission, verification of image validity, and result output. Thus, the interface developed herein includes the following features: image upload functionality, display of cropped submitted images, an authentication initiation button, and presentation of post-detection results. The detection results encompass the authenticity assessment and the likelihood of the image being authentic.

Fig.7 Fake Image Detection System